\documentclass[12pt, fceqn]{article}
\usepackage{amsfonts, amsthm, latexsym, amssymb}

\usepackage[fleqn]{amsmath}
\usepackage{graphicx}
\usepackage{graphics}
\usepackage[comma, numbers]{natbib} % for bibliography
\usepackage[comma]{natbib} % for bibliography
\usepackage{color}
\usepackage{comment}
\usepackage{algorithm}
\usepackage{algorithmic}
\usepackage{lscape}
\usepackage{footnote}
\usepackage{url}

\usepackage{epsfig}
\usepackage{fancybox}
\usepackage{psfrag}
\usepackage{tabularx}
\usepackage{float}
\usepackage{multirow}
\usepackage{rotating}
\usepackage{cancel}
\usepackage{epstopdf}
\usepackage{booktabs}

\date{}

\theoremstyle{remark}

\theoremstyle{definition}

{\begin{list}{}{\setlength{\itemsep}{4pt}
			\parindent 0pt  \setlength{\parsep}{0pt}\setlength{\leftmargin}{+25pt}%\parindent}
		\setlength{\itemindent}{-\parindent}}}{\end{list}}
\begin{document}
	
\begin{center}
	{\Large \textbf{Integrating Large Language Models with Network Optimization for Interactive and Explainable Supply Chain Planning: A Real-World Case Study}}\\[12pt]
	
	% Authors and addresses:
	
	\footnotesize
	
	\mbox{\large Saravanan Venkatachalam}\\
	
	Department of Industrial and Systems Engineering, Wayne State University, \\ 4815 Fourth St, Detroit, MI 48202.\\
	Corresponding author: \mbox{saravanan.v@wayne.edu}\\
	\normalsize
\end{center}

\begin{abstract}
This paper presents an integrated framework that combines traditional network optimization models with large language models (LLMs) to deliver interactive, explainable, and role-aware decision support for supply chain planning. The proposed system bridges the gap between complex operations research outputs and business stakeholder understanding by generating natural language summaries, contextual visualizations, and tailored key performance indicators (KPIs). The core optimization model addresses tactical inventory redistribution across a network of distribution centers for multi-period and multi-item, using a mixed-integer formulation. The technical architecture incorporates AI agents, RESTful APIs, and a dynamic user interface to support real-time interaction, configuration updates, and simulation-based insights. A case study demonstrates how the system improves planning outcomes by preventing stockouts, reducing costs, and maintaining service levels. Future extensions include integrating private LLMs, transfer learning, reinforcement learning, and Bayesian neural networks to enhance explainability, adaptability, and real-time decision-making.
\end{abstract}

% %%Graphical abstract
% \begin{graphicalabstract}
% %\includegraphics{grabs}
% \end{graphicalabstract}

%Research highlights

\section{Introduction}\label{sec-intro}

Network optimization plays a crucial role in supply chain planning by efficiently managing product flows between various locations, such as suppliers, distribution centers (DCs), and retail stores \cite{hugos2018essentials, lambert2000issues}. Typically, mathematical optimization models like linear or mixed-integer programming (MIP) are employed due to their precision, sophisticated modeling capabilities, and robust problem-solving power. However, interpreting results from these optimization models often poses significant challenges, especially for users unfamiliar with operations research (OR). Also, users with different roles within an organization would like to see the recommendations from network optimization differently. Recently, large language models (LLMs), particularly transformer-based encoders and decoders known for their powerful text-generation capabilities, have rapidly advanced and found widespread adoption across industries \cite{openai2023gpt-4, bubeck2023sparks, lee2023benefits}. This study explores leveraging LLMs to improve the explainability and user-friendliness of OR model solutions \cite{mostajabdaveh2025evaluating, li2023large}. By integrating an LLM into the OR modeling framework, complex optimization outcomes can be translated into clear, interactive summaries and explanations that cater specifically to the diverse information needs of non-technical stakeholders. This approach not only simplifies interpretation but also enhances decision-making confidence among business users. A real-world example is provided to demonstrate how effectively an LLM can serve as an explanatory layer around traditional optimization engines, improving clarity, usability, and stakeholder engagement in network planning tasks. Ultimately, this integration aims to bridge the communication gap between OR specialists and business decision-makers, fostering more informed, transparent, and collaborative planning processes.

\begin{figure}
	\centering
	\includegraphics[scale=0.4]{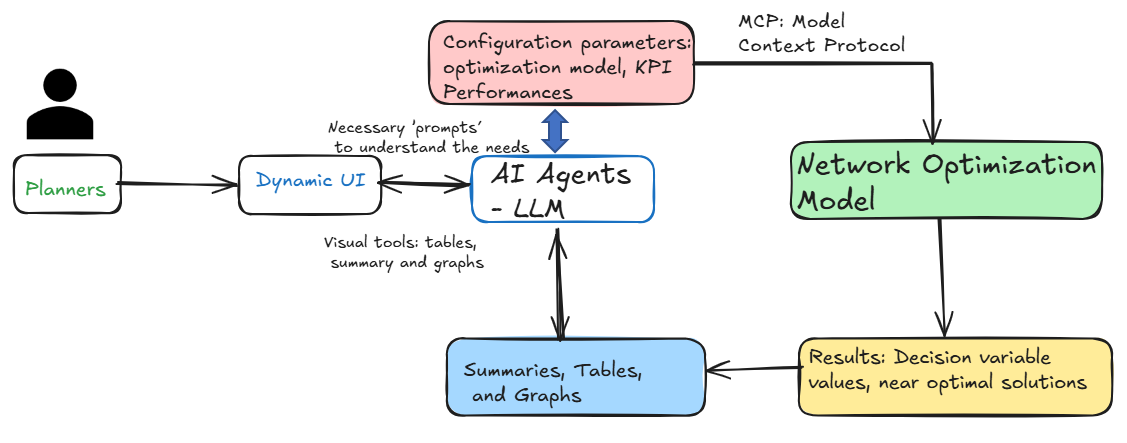}
	\caption{LLM-Driven Optimization Architecture: A modular system integrating user prompts, visual tools, AI agents, and optimization models to support decision-making through dynamic interfaces and the Model Context Protocol (MCP). Results are presented via summaries, tables, and graphs to aid interpretation and action.}
	\label{fig:overall}
\end{figure}

Figure~\ref{fig:overall} illustrates the workflow of an AI-assisted decision-support system for network optimization. Planners interact with a dynamic user interface that captures their operational needs and planning objectives. This interface communicates with AI agents, which are powered by LLMs, by sending prompts that help interpret the user’s intent. These AI agents process the inputs and integrate configuration parameters such as optimization model settings and key performance indicators. The processed information is then passed to the optimization model for network optimization using a model context protocol (MCP) \cite{ray2025survey}. The optimization engine solves the problem and returns results in the form of decision variable values and near-optimal solutions. These results are interpreted and transformed by the AI agents into summaries, tables, and graphs. The visual outputs are finally returned to the user interface, allowing planners to review and analyze the recommendations in a clear and interactive format, thus completing a closed-loop feedback system.

\section{Background and Motivation} \label{subsec:contributions}

Inventory balancing through network optimization is a cornerstone of modern supply chain planning \cite{venkatachalam5349428inventory}. It plays a pivotal role in enhancing operational efficiency, service level reliability, and cost-effectiveness across strategic, tactical, and operational decision-making levels. Strategically, network design decisions determine the optimal configuration of manufacturing plants, distribution centers (DCs), and retail nodes \cite{taghizadeh2021impact, venkatachalam2016efficient}. Tactically, these decisions govern how inventory is dynamically redistributed in response to evolving demand and supply uncertainties. Operationally, they help ensure timely sourcing and fulfillment, especially when dealing with long lead times and volatile customer behavior.

In the current landscape, characterized by globalized supply chains and increased demand volatility, companies must frequently re-evaluate how inventories are allocated across their networks \cite{taghizadeh2021impact, taghizadeh2023two, venkatachalam2019two}. This is particularly true in systems where products are sourced from distant facilities and delivered through regional hubs. In such environments, even minor disruptions or forecast errors can lead to substantial mismatches between available stock and actual demand. To counteract this, supply chain planners engage in tactical inventory rebalancing: transferring stock between DCs to proactively prevent shortages, minimize excess, and maintain service continuity. These decisions must consider a variety of constraints such as shipment minimums, cost thresholds, forecast uncertainty, and lead time constraints.

This work focuses on a real-world tactical planning scenario where a retailer operates a network of stores across the U.S., with supply aggregation occurring at regional DCs. With supply originating from offshore facilities and lead times ranging from 14 to 20 weeks, DCs must serve as adaptive buffers to absorb forecast inaccuracies and short-term demand shocks. Efficient inventory transfers between DCs, based on anticipated future needs and system-wide availability, become crucial levers in safeguarding product availability.

While optimization models such as MIP offer mathematically sound solutions for these redistribution problems, their outputs are often not readily interpretable by planners, managers, or executives. These users require actionable insights in natural language and visuals—not abstract variables or objective function values. This gap in interpretability presents a significant barrier to the broader adoption of operations research (OR) tools in practice. Furthermore, in time-sensitive scenarios, the delay in translating model outputs into operational decisions can reduce the practical utility of such models.

Traditional OR-based decision systems often assume a static decision-maker who can parse solver outputs and translate them into action. However, real-world planning environments are increasingly collaborative and fast-paced, involving multiple stakeholders with varying informational needs and technical expertise. Business users—ranging from SKU-level analysts to region-focused executives—demand intuitive interfaces that summarize outcomes in ways that align with their priorities and language. This calls for a new layer in decision systems: one that is not only computationally efficient but also cognitively aligned with human users.

In response to this need, we explore the integration of Large Language Models (LLMs) as a natural language interface to optimization engines. LLMs can interpret, contextualize, and summarize optimization results while tailoring explanations to specific user roles. Their ability to dynamically generate multi-level, query-driven content enables real-time interactivity in ways that traditional dashboard tools cannot match. By embedding LLMs into the optimization loop, we enable a form of “explainable optimization”—where outcomes are not only computed but also communicated in a transparent, responsive, and role-aware manner. This hybrid architecture redefines the role of AI in supply chain planning—not as a replacement for human decision-making, but as a facilitator of faster, clearer, and more informed decisions.

\subsection{Optimization Models} \label{subsec:optmodels}
Optimization models provide an effective way to represent and solve planning problems, particularly where decisions involve allocating limited resources optimally to meet specific business requirements \cite{garcia2015supply}. These models allow decision-makers to explicitly define operational constraints and clearly establish objectives, such as minimizing costs, maximizing profits, or improving resource utilization \cite{pourhejazy2016new}. Among the available optimization techniques, linear programming and MIP are extensively used across various industries due to their precision, versatility in modeling complex real-world scenarios, and consistent quality of solutions \cite{nickel2022decision}. Another major benefit of optimization models is that they deliver proven optimal or near-optimal solutions, thus offering confidence and reliability in the decision-making process. However, despite these advantages, formulating, solving, and interpreting MIP models usually demands specialized knowledge, which typically resides within OR practitioners \cite{cyras2019argumentation}. Although the concept of “optimality” resonates clearly within the OR community, effectively explaining and communicating these optimized solutions to business stakeholders remains challenging. The inherent complexity and technical nature of these models can result in a significant gap between the model developers and end-users who rely on these solutions to make critical business decisions. Consequently, this communication barrier often reduces the practical adoption and widespread implementation of OR models in real-world business environments. To address this critical challenge, our work focuses on developing user-friendly tools and clear explanatory methods that translate complex optimization outputs into straightforward, easily understandable insights tailored to business users. By enhancing the interpretability of optimization solutions through intuitive explanations, visualization, and interactive summaries, we aim to bridge the gap between OR specialists and decision-makers. This, in turn, promotes greater trust, understanding, and collaboration within organizations, ultimately leading to broader acceptance and more informed usage of optimization methods in supply chain planning and decision-making processes.

\subsection{Large Language Models} \label{llms}
Large language models (LLMs) are playing a significant role in the growing application of AI across various domains \cite{github2023copilot, chen2023frugalgpt}. In natural language processing, LLMs have replaced traditional statistical and rule-based methods with neural networks trained on vast amounts of text data, enabling them to capture complex linguistic patterns and relationships \cite{devlin2018bert, rosset2020turing-nlg, smith2022using}. Built on transformer architectures that combine encoders and decoders, LLMs efficiently process sequences of text based on learned context \cite{liu2023prompt, brown2020language}. They are increasingly impactful in data analysis, where natural language queries can be translated into executable code for analysis and visualization, allowing business users to perform complex tasks with ease \cite{danilevsky2020survey, ahmed2022artificial}. Moreover, LLMs excel at interpreting results by transforming the outputs of OR models—such as solution values, dual variables, and slackness conditions—into clear, coherent explanations for non-technical users. This automated data storytelling enhances understanding and supports better decision-making \cite{bommasani2021opportunities, peters2018deep}. Integrating LLMs into OR model pipelines thus improves both the execution and interpretation of results, making insights more accessible and boosting confidence among business users.

Additionally, users within an organization often have varying informational needs based on their roles. For example, analysts may be interested in insights at the item level, managers may focus on product families, while senior executives are typically concerned with performance at the regional or location level. Designing static user interfaces to cater to each of these perspectives can be both complex and inflexible. However, with the capabilities of LLMs, dynamic and role-specific descriptive summaries of key performance indicators (KPIs) can be generated on demand. This adaptability allows each user to receive insights tailored to their level of responsibility and decision-making needs. As a result, integrating OR models with LLMs creates a comprehensive and flexible pipeline that supports users across different organizational levels, improving accessibility, clarity, and overall effectiveness in decision-making.

\section{Problem Formulation} \label{sec:formulation} 
We start with a description of the network optimization model (NOM), followed by notations and then the mathematical formulation. 
\subsection{Description} \label{subsec:informal-problem-statement}
We consider a complex network optimization problem involving multiple products, time periods, and DCs, managed by a centralized supply chain planner. The planner is responsible for determining the optimal timing and quantity of inventory transfers between DCs to address potential shortfalls in retail store demand before the next scheduled replenishments arrive. This represents a tactical planning problem where the primary objective is to rebalance inventory across the network in a way that ensures service level continuity and minimizes disruptions.

For each stock keeping unit (SKU) at a DC, the planner must ensure that the transferring DC retains enough inventory and safety stocks to satisfy the upcoming demands of its own assigned retail stores while still being able to support other DCs experiencing stock shortages. The receiving DC, on the other hand, aims to reduce the risk of stockouts by accepting additional inventory through these transfers. Each inter-DC transfer incurs a minor order cost, and transfers must meet a minimum order quantity to justify the logistical effort and cost. To avoid redundancy and unnecessary movement of goods, transshipments occurring within the same time period between DCs are not allowed.

The overall objective of the optimization model is to minimize the total stockout levels and order costs while incentivizing the maintenance of sufficient safety stock for each SKU across all DCs. This must be achieved under various operational constraints such as inventory availability, minimum transfer quantities, and lead times. The problem introduces intricate interdependencies between decisions across SKUs, DCs, and time periods, significantly increasing the complexity of the solution space. Addressing this challenge requires a robust and scalable optimization approach that can balance competing goals and provide actionable plans for tactical inventory movement across the network.

\subsection{Notation} \label{subsec:notation}
\paragraph{Sets and Indices}
Let $\mathcal{P}$ denote the set of SKUs, which is an item at a distribution center, indexed by $i$, and let $\mathcal{T}$ be the set of time periods, indexed by $t$. The time horizon $\mathcal{T}$ is partitioned into two disjoint subsets, $\mathcal{T}_1$ and $\mathcal{T}_2$, such that $\mathcal{T} = \mathcal{T}_1 \cup \mathcal{T}_2$. The subset $\mathcal{T}_1$ represents the `frozen periods', during which inventory transfers between distribution centers (DCs) are not allowed, as these periods are too close to the actual demand to permit feasible adjustments. In contrast, $\mathcal{T}_2$ comprises the `transfer-eligible periods', where inter-DC transfers are permitted.

\begin{comment}
et $\mathcal{I}$ be the set of products or items, indexed by $i$, and $\mathcal{T}$ the set of time periods, indexed by $t$. The set of freight discount breakpoints is denoted by $\mathcal{B}$, indexed by $b$. Additionally, the product set $\mathcal{N}$ is partitioned into $k$ disjoint subsets, denoted as $\mathcal{N}_1 \cup \dots \cup \mathcal{N}_k$. Here, $\mathcal{N}_j \subseteq \mathcal{N}$ for any $j \in \{1, \dots, k\}$ denotes a subset of items for which a certain degree of proportional inventory balance must be maintained at all times. In the vanilla formulation of the CCRP, as used in the literature \cite{ertogral2008multi,venkatachalam2015efficient,venkatachalam2019two,taghizadeh2023two}, and presented here first, no such balance constraints are imposed. In later parts of this section, we introduce a modified formulation called the $\varepsilon$-CCRP, where an $\varepsilon$-proportional balance condition presented in Sec.~\ref{sec:inventory-balancing}, is enforced across all items in each subset \( \mathcal{N}_j \).
\end{comment}

\paragraph{Parameters and Decision Variables}
The presented formulation addresses a multi-period inventory and transshipment optimization problem involving a set of DCs and SKUs, denoted by $\mathcal{P}$, over a discrete planning horizon $\mathcal{T}$. The objective is to maximize the total net benefit derived from maintaining adequate safety stock levels, while minimizing penalties associated with unmet demand and costs incurred from inter-DC shipments. The key parameters include the safety stock benefit per unit, $\hat{h}_{it}$, representing the value of meeting demand using available inventory for SKU $i$ in period $t$; the shortage penalty $\hat{k}_{it}$ for any unsatisfied demand; and the fixed shipment cost $\hat{r}_{it}$ incurred when a minimum quantity is shipped to a DC. Additional parameters include the demand $\hat{d}_{it}$, the SKU-specific safety stock level $\hat{s}_{it}$, the initial inventory $\hat{I}_{i0}$, a sufficiently large constant $\hat{M}$ used in big-M constraints, and the minimum shipment threshold $\hat{Q}$.

The decision variables capture planning decisions over time: $X_{ii't}$ denotes the quantity of SKU shipped from DC $i$ to DC $i'$ in period $t$; $I_{it}$ represents the net inventory at DC $i$ and can take both positive and negative values; $IP_{it}$ is the positive component of $I_{it}$, accounting for safety stock and excess inventory; and $IM_{it}$ is the negative component, representing the shortfall due to unmet demand. Binary variables $Y_{it}$ and $Z_{it}$ respectively indicate whether a DC is active (i.e., eligible to receive shipments) and whether the minimum shipment condition is enforced in period $t$.

\subsection{Network Optimization Model Formulation} \label{subsec:ccrp}
We present the network optimization model where the constraints ensure operational feasibility by maintaining inventory balance across time periods, restricting shipments to only active DCs, and preventing reciprocal transshipments between DCs in the same period. Minimum shipment constraints are included to reflect realistic logistics requirements, and inventory is decomposed into positive and negative components to distinguish between safety stock, excess, and unmet demand.

Frozen period constraints prevent last-minute changes, aligning the model with practical lead-time considerations. All decision variables are bounded appropriately—binary for activation flags, continuous for inventory levels, and non-negative for shipment quantities—ensuring both mathematical consistency and real-world applicability. Together, these constraints embed key business rules into the model, enabling the generation of feasible, cost-effective, and implementable plans addressing the operational restrictions for the company.

\begin{subequations}
\begin{align}
\text{Objective function}: \quad &\max ~~\sum_{t \in \mathcal{T}} \left[ \sum_{i \in \mathcal{P}} \left( \hat{h}_{it} IS_{it} - \hat{k}_{it} IM_{it} - \hat{r}_{it} Z_{it} \right)  \right] \label{eq:obj} \\[0.2cm]
\text{Inventory balance for } t > 1: \quad & I_{it} = I_{i,t-1} + \sum_{i^{`} \in \mathcal{P}} X_{i^{`}it} - \hat{d}_{it} \quad \forall i \in \mathcal{P}, \; t \in |\mathcal T_2| \label{eq:ib} \\[0.2cm]
\text{Inventory balance for } t = 1: \quad & I_{i1} = \hat{I}_{i0} + \sum_{i^{`} \in \mathcal{P}} X_{i^{`}i1} - \hat{d}_{i1} \quad \forall i \in \mathcal{P} \label{eq:ib-1} \\[0.2cm]
\text{Setup enforcement:} \quad & X_{ii^{`}t} \leqslant \hat M Y_{i^{`}t} \quad \forall i, i^{`} \in \mathcal{P}, \; t \in \mathcal{T} \label{eq:setup} \\[0.2cm]
\text{No transshipment:} \quad & X_{i^{`}it} + \hat M y_{i^{`}t} \leqslant \hat M \quad \forall i, i^{`} \in \mathcal{P}, \; t \in \mathcal{T} \label{eq:notrans} \\[0.2cm]
\text{Inventory break-up:} \quad & I_{it} = IP_{it} - IM_{it}  \quad \forall i, i^{`} \in \mathcal{P}, \; t \in \mathcal{T} \label{eq:inv} \\[0.2cm]
\text{SS Break-up:} \quad & IP_{it} = IS_{it} + IE_{it}  \quad \forall i, i^{`} \in \mathcal{P}, \; t \in \mathcal{T} \label{eq:ssbreak} \\[0.2cm]
\text{Min quantity enforcement:} \quad & \sum_{i^`\in \mathcal{P}} X_{ii^{`}t} \leqslant \hat M Z_{it} \quad \forall i \in \mathcal{P}, \; t \in \mathcal{T} \label{eq:minqtyen} \\[0.2cm]
\text{Min quantity:} \quad & X_{i^{`}it} \geqslant \hat Q - (1-Z_{i^`t}) \hat Q \quad \forall i, i^{`} \in \mathcal{P}, \; t \in \mathcal{T} \label{eq:minqty} \\[0.2cm]
\text{Frozen period:} \quad & X_{i^{`}it} \leqslant 0 \quad \forall i, i^{`} \in \mathcal{P}, \; t \in \mathcal{T}_{1} \label{eq:froper} \\[0.2cm]
\text{Safety stock limit:} \quad & IS_{it} \leqslant \hat{s}_{it} \quad \forall i, i^{`} \in \mathcal{P}, \; t \in \mathcal{T} \label{eq:ss} \\[0.2cm]
\text{Variable domains:} \quad & Y_{it}, Z_{it} \in \{0,1\}, \quad I_{it} \in \mathbb{R}, \quad X_{ii^{`}t}, IP_{it}, IS_{it}, IE_{it}, IM_{it} \geqslant 0 \label{eq:var-domains}
\end{align}
\end{subequations}

The mathematical formulation includes several constraints that ensure the feasibility and logic of the inventory and transshipment planning model. The objective function~\eqref{eq:obj} maximizes the overall net benefit by accounting for the value gained from satisfying uncertain demand using safety stock, penalizing unmet demand, and deducting the cost associated with triggering minimum shipment quantities. The inventory balance is maintained across periods through constraints~\eqref{eq:ib} and~\eqref{eq:ib-1}, where the former handles periods after the first by linking current inventory to previous inventory, incoming shipments, and demand, while the latter uses the initial inventory in place of past inventory for the first period. The setup enforcement constraint~\eqref{eq:setup} ensures that a DC can only receive shipments if it is active during that period, while the no transshipment constraint~\eqref{eq:notrans} prevents reciprocal or looping shipments between DCs in the same period.

The inventory decomposition constraint~\eqref{eq:inv} splits net inventory into positive and negative components to distinguish between excess stock and shortages, and the safety stock decomposition in~\eqref{eq:ssbreak} further breaks the positive component into the portion used to meet safety stock, and the remaining excess. To model shipment thresholds, constraint~\eqref{eq:minqtyen} restricts outbound shipments based on whether the minimum shipment flag is activated, and~\eqref{eq:minqty} enforces the actual minimum quantity requirement when the flag is active. The frozen period constraint~\eqref{eq:froper} prohibits any shipments during designated periods where decisions are locked in advance. The safety stock limit~\eqref{eq:ss} ensures that the amount of inventory used to fulfill `uncertain' demand does not exceed the defined safety stock level for each SKU and period. Finally, the variable domain constraint~\eqref{eq:var-domains} defines the valid ranges and types of all decision variables, including binary, continuous, and non-negative domains as appropriate.

\section{Technical Architecture} \label{ta}

\begin{figure}[H]
	\centering
	\includegraphics[scale=0.5]{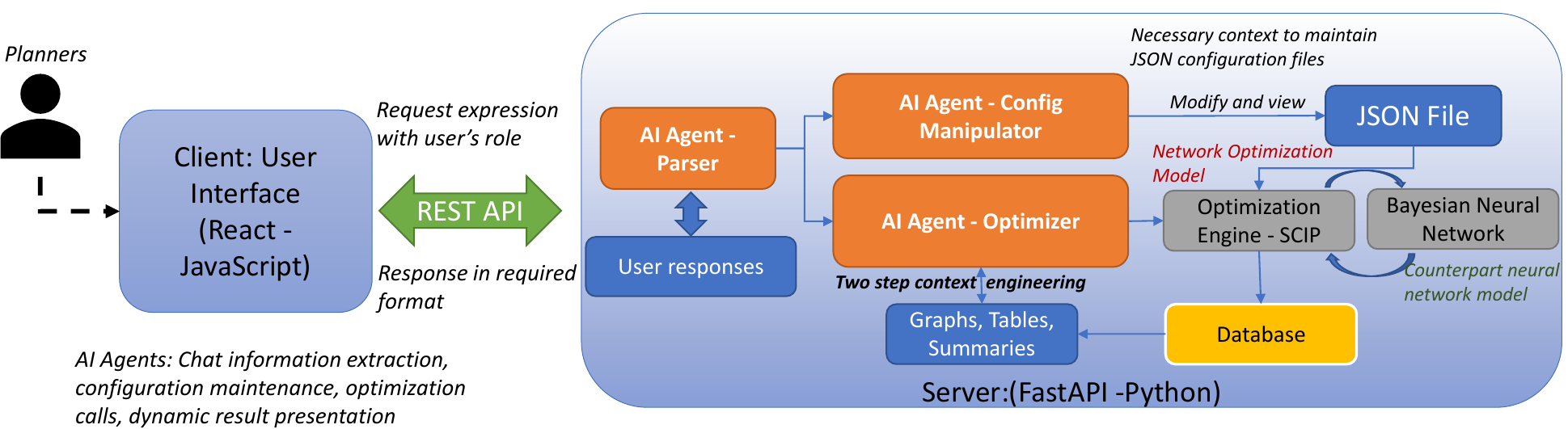}
	\caption{System architecture for interactive and explainable supply chain planning, integrating a role-aware user interface with AI agents, optimization engines, and dynamic result generation.}
	\label{fig:sysarch}
\end{figure}

The illustrated architecture in Figure \ref{fig:sysarch} presents an AI-powered decision-support system tailored for network optimization in supply chain planning. It is designed to support planners across various roles by integrating a user-friendly interface with a powerful backend system. The backend consists of AI agents, an optimization engine, a neural network model, and a dynamic result interpretation layer. The system emphasizes interactivity, role-aware recommendations, and explainability—bridging the gap between complex operations research models and real-world decision-making.

\subsection{User Interaction and Role-Aware Interface}

At the core of the system is a web-based interface developed using React and JavaScript, through which planners across different organizational roles interact with the system. This interface enables users to pose queries or submit planning requests in natural or structured language. These requests may vary greatly based on user roles: for instance, analysts often need granular insights at the SKU or item level, managers may be interested in product families or category-level performance, while executives usually prefer summarized KPIs at a location or regional level.

This role-awareness is critical, as it allows the system to tailor its responses in both format and detail. For example, a simple request like ``Show projected stockouts'' might trigger different levels of data aggregation and explanation depending on whether the requester is an analyst or a senior leader. Once submitted, user requests are sent to the server via a REST API, ensuring seamless integration between the front-end and the back-end. The interface also serves as a visualization hub, where results are presented in clear formats such as charts, tables, or narrative summaries that align with the user’s responsibilities.

\subsection{AI Agents and Request Interpretation}

Once a request reaches the backend (built with FastAPI in Python), a series of AI agents come into play to interpret and process it. The first of these is the `AI Parser Agent', which reads the incoming request and extracts relevant components—such as product identifiers, time windows, types of decisions (e.g., transfer planning, stockout minimization), change the configuration settings, run the optimization model, and user context. This structured parsing transforms ambiguous natural language inputs into machine-readable instructions.

Following the parsing stage, if the request is to view or modify the configuration settings, the `AI Config Manipulator Agent' manages the system's configuration layer, which is stored in flexible, human-readable JSON files. These files contain the core parameters for the optimization models, including inventory policies, supply constraints, DC relationships, service levels, and demand forecasts. The Config Manipulator validates, updates, or retrieves these configuration details as required, making the system adaptable and customizable without hard-coded changes. This modular configuration setup ensures that planners can run various what-if scenarios or reconfigure planning assumptions easily, improving usability and flexibility.

\subsection{Optimization and Machine Learning Integration}

When a parsed request corresponds to an optimization run, the AI Optimizer Agent is activated. It performs two key tasks: first, it builds the optimization model by engineering the input context using parsed queries and configuration data; second, it interprets the solver’s output, translating it into clear, decision-ready formats. This two-step context engineering process ensures that both the model inputs and outputs are meaningful, relevant, and aligned with the planner’s needs.

The core optimization is handled by SCIP \cite{schwarz2010introduction}, a robust and high-performance solver for mixed-integer programming problems. SCIP obtains optimal or near-optimal solutions for complex planning decisions, such as how much inventory to transfer between DCs, when to ship, and how to manage safety stocks and avoid stockouts. The model considers various operational constraints such as lead times, minimum order quantities, transfer windows, and frozen periods where changes are not allowed.

In parallel, the system can also leverage a Bayesian Neural Network (BNN)—a machine learning model that provides probabilistic predictions and quick approximations. This model is useful for generating insights when rapid responses are needed or when full optimization runs are computationally intensive. The BNN can also support learning from past decisions stored in the system, improving performance over time. Together, SCIP and the BNN create a hybrid optimization-intelligence framework that blends mathematical rigor with learning-driven adaptability.

\subsection{Data Flow, Output Generation, and Result Presentation}

After the optimization or inference process is complete, the system moves into result generation and presentation. All outputs—whether generated by SCIP, the Bayesian Neural Network, or precomputed data—are stored in a centralized database. This database serves not only as a result repository but also as a memory bank that supports learning, auditing, and historical comparison. It ensures that planners can trace decisions, monitor trends, and assess changes over time.

The final outputs are processed by the Optimizer Agent into intuitive visualizations and summaries. These include graphs (e.g., inventory trends, transfer flows), tables (e.g., DC-level shortages, order quantities), and natural-language summaries that explain what the data means in plain English. The use of large language models (LLMs) in this layer enhances accessibility by converting complex numerical and structural outputs into coherent, role-specific narratives. These summaries help users at all levels—technical or not—to understand the implications of the optimization outcomes, identify actionable insights, and make confident decisions. The result is a fully interactive, explainable, and adaptive planning system that aligns advanced analytics with real-world business needs.

\section{Implementation} \label{sec:algorithms}

This section describes the end-to-end implementation of the network optimization and context engineering system. The goal is to deliver explainable, data-driven insights to supply chain planners and decision-makers. The implementation integrates a dynamic optimization dashboard, real-time transfer visualization, site-level inventory tracking, and a robust context engineering framework powered by LLMs. The system architecture supports operational responsiveness and explainability through an interactive user interface and intelligent backend reasoning.

\subsection{Dashboard}

The \textit{Network Optimization Dashboard} serves as the central interface for monitoring and managing supply chain network performance. It displays key system metrics including the number of active nodes, an optimization score, and projected cost savings, reflecting high operational efficiency. The dashboard allows users to execute optimization runs, configure model parameters, and input contextual data. Quick-action buttons streamline essential operations such as tolerance updates and configuration retrieval. Tab-based navigation enables access to modules related to network configuration, transfers, supply-demand analysis, and optimization results. With full node connectivity, the dashboard ensures transparent, real-time control across the network.

\begin{figure}[H]
	\centering
	\includegraphics[width=\textwidth]{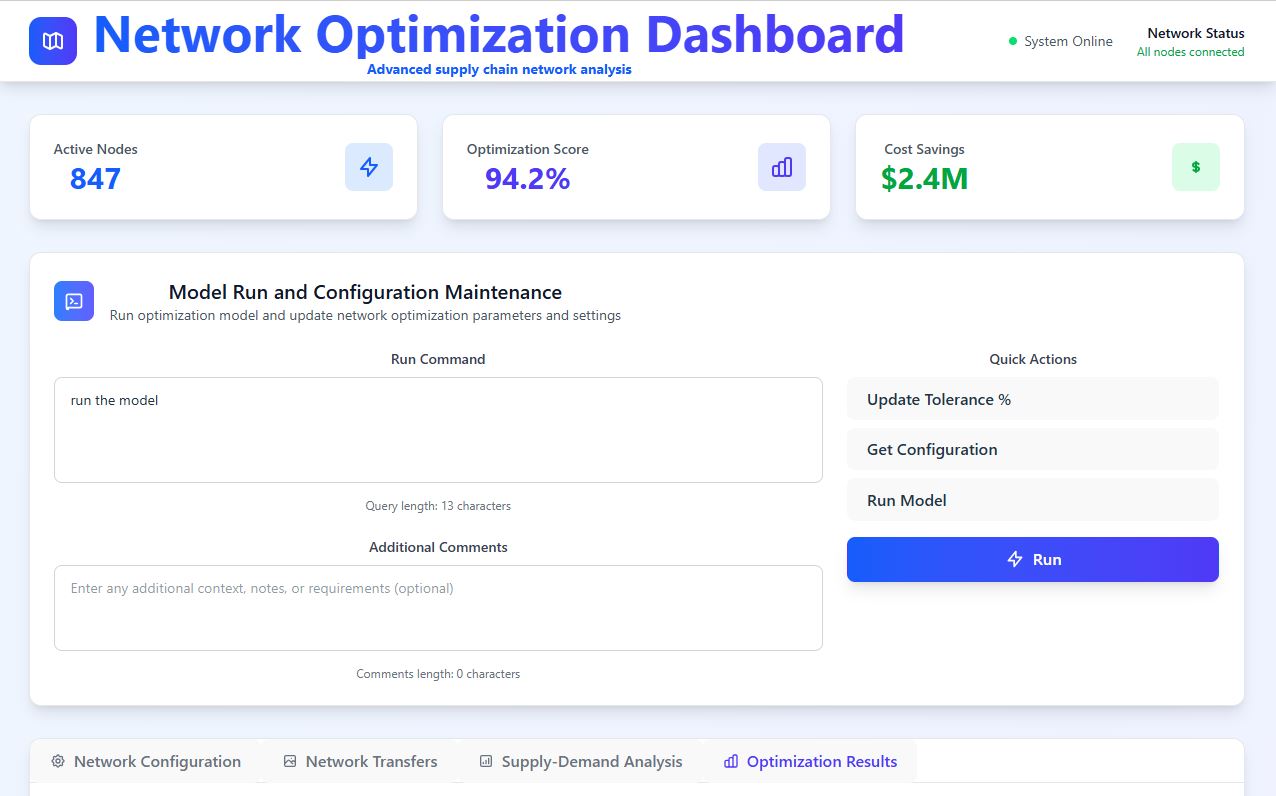}
	\caption{Network Optimization Dashboard interface displaying real-time metrics on active nodes, optimization score, and cost savings. The dashboard enables users to execute optimization models, adjust configuration parameters, and access various modules related to network performance and supply chain analysis.}
	\label{fig:dash}
\end{figure}

\subsection{Context Engineering Framework}

\begin{figure}[H]
	\centering
	\includegraphics[width=\textwidth]{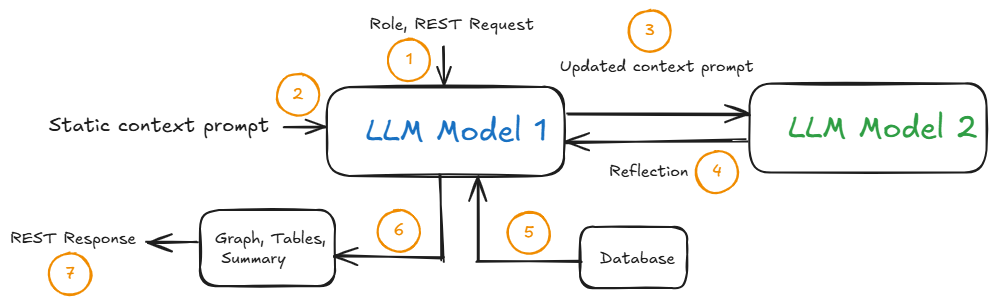}
	\caption{Context engineering architecture integrating LLMs, REST APIs, and data systems to deliver interactive, explainable, and role-specific decision support in seven structured steps.}
	\label{fig:arch}
\end{figure}

To enable explainable and role-specific outputs, the system employs a context engineering (CE) pipeline powered by large language models (LLMs), structured data, and REST APIs. Figure \ref{fig:arch} illustrates this process, which unfolds across seven key steps.

The pipeline begins with Step 1, where LLM Model 1 receives inputs from the user interface, including the user’s role and the REST request, alongside a static CE template. This template includes foundational information such as few-shot prompt examples, optimization model constraints and variables, KPI definitions (e.g., weeks of supply (WOS), cost), rationale for inter-DC transfers, and structured metadata for both input and output. In Step 2, LLM Model 1 dynamically modifies this static CE template based on the user role and specific request, using a prompt templating approach to ensure that the context is personalized and relevant.
The updated context is passed to LLM Model 2 in Step 3, which initiates a reflection mechanism in Step 4. This mechanism is used to assess and verify the completeness and quality of the engineered context, ensuring that it meets the informational and operational needs of the optimization pipeline. Notably, the reflection process leverages differences between LLM1 and LLM2 to enhance contextual accuracy, consistency, and modularity. In Steps 5 and 6, the refined context is used to query backend systems for relevant structured data. This data is then transformed into user-specific outputs such as tables, graphs, summaries, and natural language explanations, tailored to the planner’s role and the optimization task. Finally, in Step 7, the generated content is packaged and returned via a REST response, completing the pipeline.

This multi-model, reflection-enabled architecture ensures that both the construction and delivery of optimization-related insights are interpretable, accurate, and aligned with user intent.

\subsection{Network Transfers}

\begin{figure}[H]
	\centering
	\includegraphics[width=\textwidth]{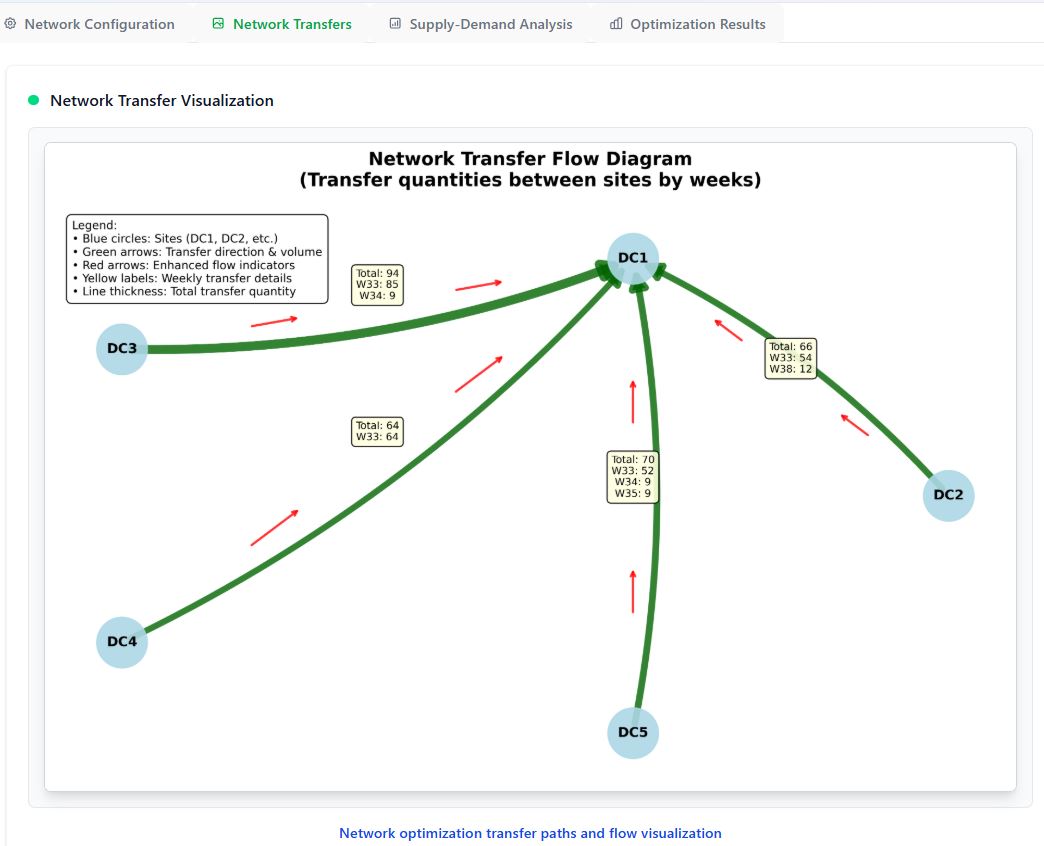}
	\caption{Network Transfer Flow Diagram visualizing inter-site transfer quantities across multiple weeks. Node labels (DC1–DC5) represent distribution centers, with green arrows indicating direction and volume of transfers, red arrows enhancing flow visibility, and line thickness corresponding to total transfer volume. Weekly quantities are annotated for detailed temporal analysis of the supply chain network.}
	\label{fig:network}
\end{figure}

Building on the dashboard’s capabilities, the system provides an intuitive visualization of inter-site transfers to support tactical decision-making. In this example, five distribution centers (DC1–DC5) are modeled. Stockouts are intentionally imposed at DC1 across all time periods, while DC2 through DC5 attempt to fulfill the unmet demand to the extent possible.  Figure~\ref{fig:network} illustrates a transfer flow diagram across distribution centers (DC1–DC5), where DC1 functions as the central redistribution hub. Green arrows denote the direction and volume of transfers, while red arrows highlight key transfer routes. Arrow thickness reflects cumulative quantity, and yellow labels annotate total and weekly breakdowns (e.g., W33, W34). This graphical layout aids rapid assessment of high-volume flows, bottlenecks, and temporal trends.

Beyond visual clarity, the network transfer diagram helps validate whether recommended transfers align with forecasted imbalances, while identifying underutilized or overburdened nodes. Weekly breakdowns support pattern recognition—helping planners evaluate the timing and criticality of interventions.

The visualization also improves stakeholder understanding by connecting model recommendations to operational intuition. For example, a spike in inbound transfers may prompt review of forecast assumptions or upstream supply issues. Combined with KPI overlays, the diagram functions as both a monitoring tool and an analytical asset for evaluating network performance.

\subsection{Demand-Supply Analysis}

\begin{figure}[H]
	\centering
	\includegraphics[width=\textwidth]{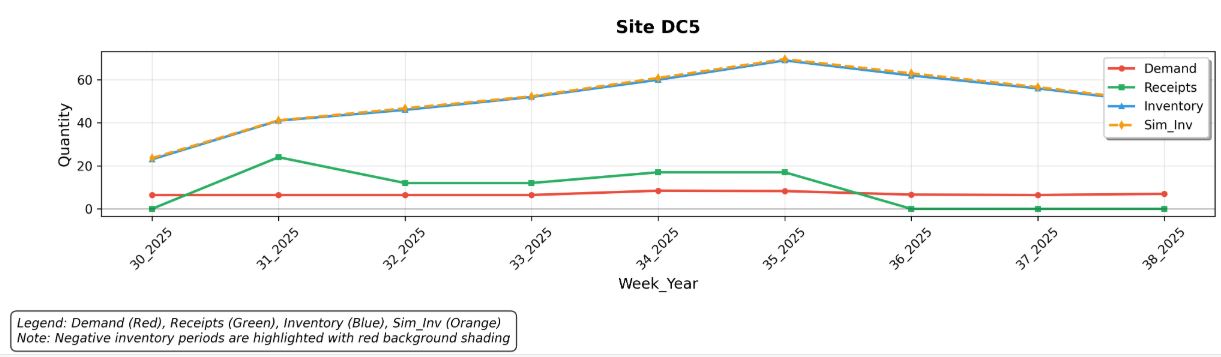}
	\caption{Weekly supply-demand and inventory dynamics at Site DC5 from Week 30 to Week 38 of 2025. The plot illustrates demand (red), receipts (green), actual inventory (blue), and simulated inventory (orange), with close alignment between actual and simulated inventories indicating accurate forecast modeling. Inventory levels peak in Week 35 before declining, driven by reduced receipts. No negative inventory periods are observed during this timeframe.}
	\label{fig:dmd}
\end{figure}

Complementing the transfer flows, site-level demand-supply dynamics are monitored to ensure inventory sufficiency and planning accuracy. Figure~\ref{fig:dmd} shows the time-series analysis at DC5 from Week 30 to 38, charting demand (red), receipts (green), actual inventory (blue), and simulated inventory (orange). Demand remains stable, while receipts spike in Week 31 and taper off toward Week 36. Inventory peaks in Week 35 before declining due to lower receipts. The alignment between actual and simulated inventory validates the forecasting model. The absence of red-shaded zones indicates that no negative inventory events occurred during this period. This analysis supports accurate forecasting, replenishment timing, and stockout risk mitigation.

\subsection{Model Execution Insights}

The integrated system was tested in a scenario involving a projected stockout at DC1 as mentioned earlier. A simulation is used to project inventories and stockouts across all DCs in the absence of a network optimization model. All the summaries presented below are generated by LLMs, leveraging the outputs from the optimization model and the CE pipeline. The CE template defines the responsibilities of a data analyst focused on assessing the effectiveness of inventory transfers within a multi-site distribution network. The analysis is driven by structured data fields including \texttt{Source\_Site}, \texttt{Destination\_Site}, \texttt{sim\_Inv} (projected inventory before transfer), \texttt{Inventory} (post-transfer inventory), \texttt{Transfer\_In}, \texttt{Transfer\_Out}, \texttt{InvCost}, \texttt{sim\_InvCost}, \texttt{Sim\_WOS}, \texttt{WOS}, \texttt{Demand}, and \texttt{Forecast}. The analyst is tasked with generating a report structured in three main sections. 

The first section, \emph{Transfer Rationale}, identifies destination sites with projected stockouts, explains which source sites provided inventory through transfers, and describes how these transfers resolved the stockouts. It also connects the stockout conditions to elevated demand or forecast values. The second section, \emph{Cost \& Performance Analysis}, explains how cost savings are achieved by replacing high stockout penalties (captured by \texttt{sim\_InvCost}) with standard holding costs (captured by \texttt{InvCost}). This section includes overall metrics such as total quantity transferred and total cost savings, along with a weekly summary table showing inventory moved and savings per week. The third section, \emph{Weeks of Supply (WOS) Impact}, evaluates how transfers impact inventory health by comparing pre- and post-transfer WOS levels at both the source and destination sites. It highlights how WOS reductions at source sites and increases at destination sites contribute to a more balanced and resilient network. The CE template enforces specific rules for formatting and analysis: only weeks with non-zero transfers are considered, and cost savings are reported only when \texttt{sim\_InvCost} exceeds \texttt{InvCost}. All field names are to be clearly highlighted in the report for readability. This structured approach ensures that inventory decisions are both explainable and actionable.

\begin{figure}[H]
	\centering
	\includegraphics[width=\textwidth]{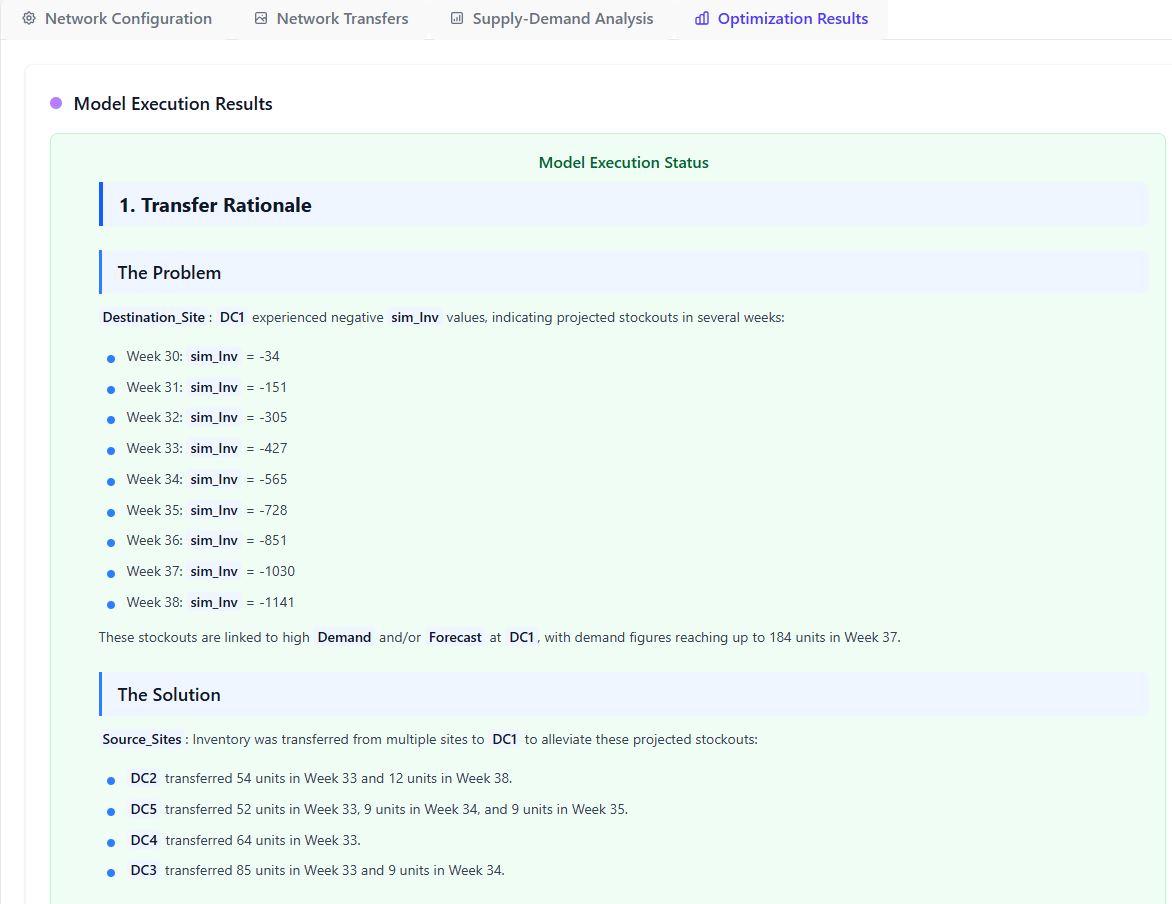}
	\caption{Model Execution Status outlining the problem of projected stockouts at DC1 due to sustained negative simulated inventory levels. The solution involved transferring inventory from multiple source sites to stabilize DC1’s stock position.}
	\label{fig:summary1}
\end{figure}

As shown in Figure~\ref{fig:summary1}, simulated inventory levels (\texttt{sim\_Inv}) at DC1 declined sharply over time, reaching a low of -1,141 units by Week 38. This significant projected stockout was primarily driven by elevated demand and forecast values, with Week 37 alone seeing demand as high as 184 units. To prevent service disruptions and maintain customer availability, inventory was proactively redistributed from upstream distribution centers (DC2–DC5). The majority of these transfers were concentrated in Week 33, during which 255 units were reallocated to address the most severe deficits.

This targeted transfer strategy not only reversed the negative inventory trajectory at DC1 but also ensured that WOS at the contributing sites remained within healthy operational limits. The use of multiple source sites enabled a balanced load-sharing approach, minimizing risk at any single location. As a result, DC1 transitioned from sustained stockouts to a stable post-transfer inventory position, enabling it to meet demand without excessive overstocking or added urgency.

\begin{figure}[H]
	\centering
	\includegraphics[width=\textwidth]{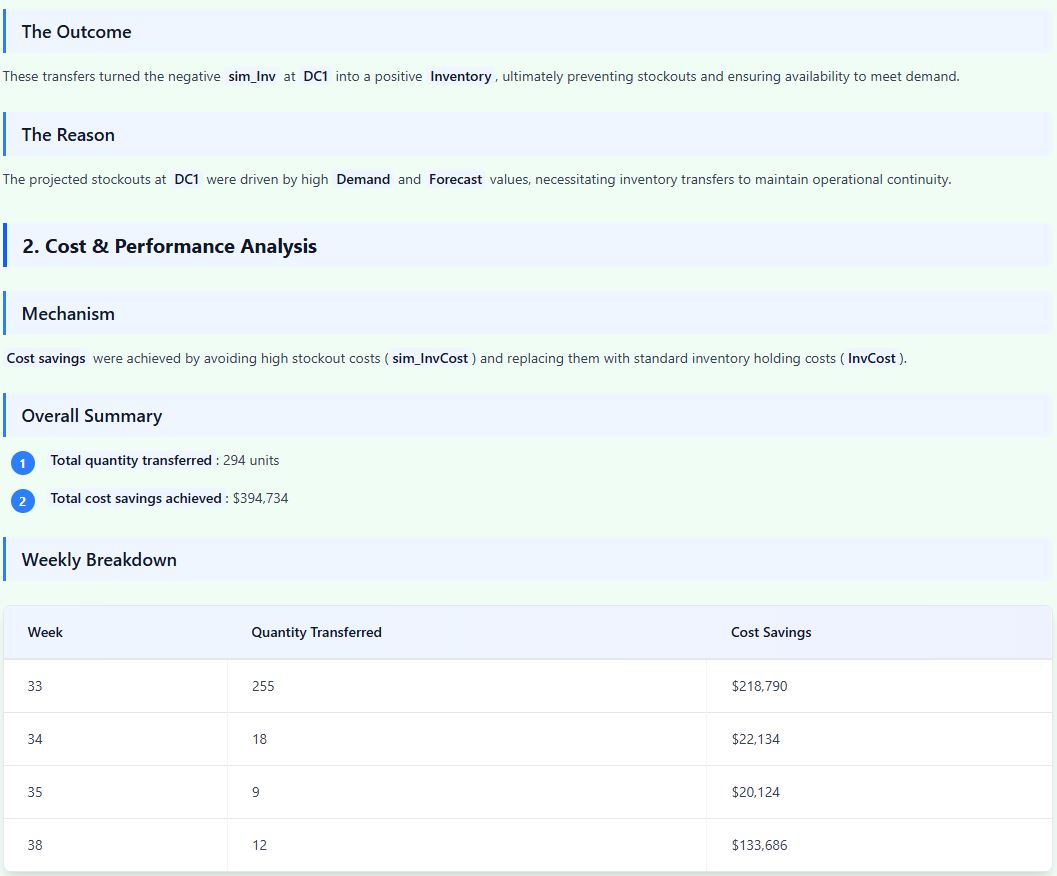}
	\caption{Cost and performance analysis of the inventory rebalancing strategy. A total of 294 units were transferred, resulting in \$394,734 in cost savings by avoiding stockout costs and using standard inventory holding costs instead.}
	\label{fig:summary2}
\end{figure}

In terms of financial impact, this approach yielded substantial efficiency gains. As illustrated in Figure~\ref{fig:summary2}, a total of 294 units were transferred across Weeks 33 to 38, leading to total cost savings of \$394,734. These savings were realized by replacing high simulated stockout penalties (\texttt{sim\_InvCost}) with regular inventory holding costs (\texttt{InvCost}), demonstrating the operational and economic value of intelligent inventory reallocation supported by network-wide visibility and optimization tools.

\begin{figure}[H]
	\centering
	\includegraphics[width=\textwidth]{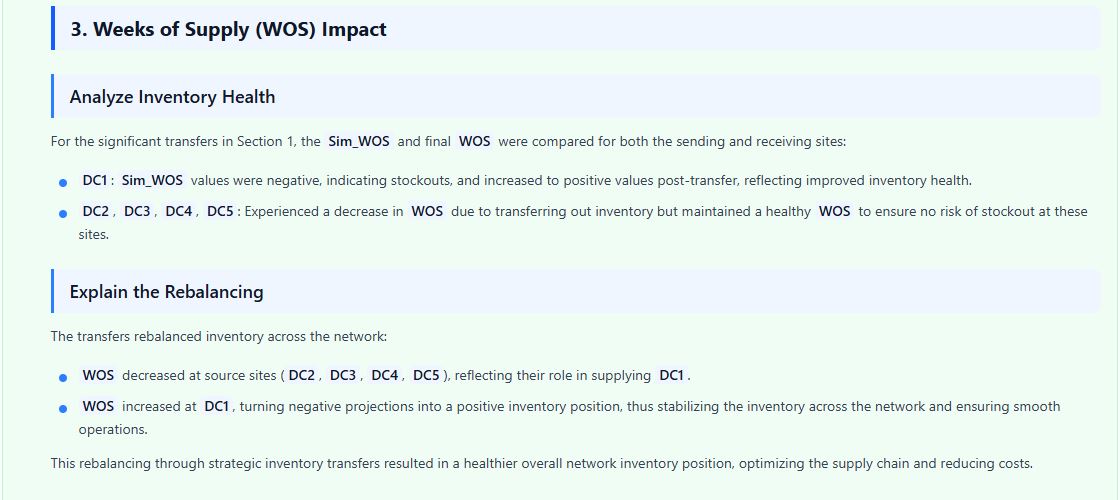}
	\caption{Weeks of Supply (WOS) impact analysis showing how post-transfer inventory levels improved at DC1 while maintaining healthy WOS at the source sites. The rebalancing helped avoid stockouts and support demand fulfillment.}
	\label{fig:summary3}
\end{figure}

As illustrated in Figure~\ref{fig:summary3}, the WOS metrics demonstrate the effectiveness of the transfer strategy in stabilizing inventory across the network. Prior to transfers, \texttt{Sim\_WOS} values at DC1 were negative or near-zero, indicating a high risk of stockouts. After the transfers, DC1’s final \texttt{WOS} increased to positive levels, reflecting a recovery in inventory health and ensuring service continuity.

Meanwhile, contributing source sites—DC2, DC3, DC4, and DC5—experienced a modest decrease in their WOS values due to outbound transfers. However, these reductions remained within acceptable thresholds, ensuring no stockouts or service degradation at the sending locations. This outcome underscores the system’s ability to redistribute inventory without compromising the stability of the overall network.

By strategically rebalancing supply across sites, the transfer mechanism transformed localized risk at DC1 into a network-wide gain. The proactive use of simulation-informed context and data-driven transfer logic enabled the system to anticipate imbalances and take corrective action in advance. This not only stabilized supply at the point of risk but also optimized inventory levels throughout the network, thereby supporting cost-effective and resilient operations.

This implementation demonstrates the integration of optimization models, LLM-based context engineering, and interactive dashboards to support intelligent supply chain decision-making. From visualizing transfer flows and site-level inventory trends to explaining complex model behavior, the system delivers explainable, accurate, and role-specific insights. Reflection-based LLM coordination further enhances context precision. Through simulation, the framework successfully prevented stockouts, optimized cost, and maintained supply continuity, showcasing its practical utility in modern operations environments.

\label{sec:comp} 
\section{Conclusion and Future Directions}\label{sec:con} 

This study presents an integrated framework that combines traditional operations research models with large language models (LLMs) to deliver interactive, explainable, and role-specific decision support for supply chain network optimization. Through a user-centric architecture that leverages AI agents, RESTful APIs, and real-time dashboards, the system bridges the gap between complex optimization logic and business stakeholder understanding. The inclusion of reflection-based context engineering further enhances the system's ability to generate coherent, accurate, and personalized insights. Simulation studies on tactical inventory redistribution demonstrated the system’s effectiveness in mitigating projected stockouts, optimizing costs, and preserving service levels across the network. 

Looking ahead, several promising avenues for future research and system enhancement emerge. First, deploying `private LLMs' \cite{touvron2023llama} trained on proprietary supply chain data can ensure stronger data privacy, security, and customization—particularly important for industry-grade deployment. In addition, `transfer learning' \cite{weiss2016survey} techniques can be employed to fine-tune general-purpose LLMs \cite{lester2021power,dettmers2023qlora} for domain-specific terminology and decision contexts, improving both the precision and relevance of generated summaries and explanations. To further improve the system’s adaptability and automation, incorporating `reinforcement learning (RL)' methods could allow the optimization engine to iteratively learn effective policies for inventory transfers under uncertainty, using historical feedback to improve future decisions. On the modeling front, leveraging a `Bayesian Neural Network (BNN)' offers a promising direction for `online network optimization', providing probabilistic predictions and uncertainty quantification that are particularly valuable in volatile supply environments. Such integration of BNNs with deterministic solvers could support rapid decision-making while maintaining robustness in dynamic and data-scarce scenarios. Finally, enhancing explainability remains an ongoing priority. Future work will focus on developing `multi-modal explanation layers' that combine textual reasoning, visual analytics, and causal attribution—allowing users to not only observe system recommendations but also understand their underlying rationale across temporal and spatial dimensions. Together, these innovations promise to elevate the role of AI-assisted tools in supply chain planning, making them more transparent, adaptive, and effective for real-world operational contexts.

\bibliographystyle{elsarticle-num} 
\bibliography{asd}
\end{document}